\documentclass[12pt]{article}
\usepackage{arxiv} 

\usepackage{amsmath}
\usepackage{amssymb}
\usepackage{times}
\usepackage{bm}
\usepackage{graphicx}

\makeatletter
\let\saved@includegraphics\includegraphics
\AtBeginDocument{\let\includegraphics\saved@includegraphics}
\renewenvironment*{figure}{\@float{figure}}{\end@float}
\makeatother

\usepackage{xcolor}

\usepackage{verbatim} 

\newenvironment{sciabstract}{%
\begin{quote} \bf}
{\end{quote}}

\title{Robust priors for regularized regression} 

\author
{Sebastian Bobadilla-Suarez,$^{1\ast}$ Matt Jones,$^{2}$ Bradley C. Love$^{1,3}$\\
\\
\normalsize{$^{1}$Department of Experimental Psychology, University College London}\\
\normalsize{26 Bedford Way, London, WC1H 0AP, UK}\\
\normalsize{$^{2}$Psychology and Neuroscience, University of Colorado Boulder}\\
\normalsize{Boulder, CO 80309-0345, USA}\\
\normalsize{$^{3}$The Alan Turing Institute, 96 Euston Road, London, NW1 2DB, UK}\\
\normalsize{$^\ast$To whom correspondence should be addressed; E-mail:  sebastian.suarez.12@ucl.ac.uk.}\\
\\
0000-0002-5951-0772 (SBS) \\ 0000-0002-4469-7896 (MJ) \\ 0000-0002-7883-7076 (BCL)
}

\date{}

\newcommand{\argmax}{\mathop{\mathrm{arg\,max}}}
\newcommand{\argmin}{\mathop{\mathrm{arg\,min}}}
\newcommand{\sign}{\mathop{\mathrm{sign}}}

\begin{document} 


\baselineskip24pt


\maketitle 

\begin{sciabstract}
Induction benefits from useful priors. Penalized regression approaches, like ridge regression, shrink weights toward zero but zero association is usually not a sensible prior. Inspired by simple and robust decision heuristics humans use, we constructed non-zero priors for penalized regression models that provide robust and interpretable solutions across several tasks. Our approach enables estimates from a constrained model to serve as a prior for a more general model, yielding a principled way to interpolate between models of differing complexity. We successfully applied this approach to a number of decision and classification problems, as well as analyzing simulated brain imaging data. Models with robust priors had excellent worst-case performance. Solutions followed from the form of the heuristic that was used to derive the prior. These new algorithms can serve applications in data analysis and machine learning, as well as help in understanding how people transition from novice to expert performance.
\end{sciabstract}





\clearpage


\textbf{Summary:} One challenge in statistical inference is choosing a robust prior that performs well across environments. We find that priors inspired by human decision-making heuristics lead to robust and interpretable models for data analysis. Heuristics are simple and ignore covariance amongst predictors. Nevertheless, they are robust, often outperforming complex models that utilize more information. Our approach integrates such priors with regularized regression in a principled way across several domains. Much like people transition from novice (heuristic-based) to expert (exact weighting) performance, models with heuristic priors are effective in applications, like analyzing neuroimaging data, in low and high data availability regimes.



\section{Introduction}

Inference from data is most successful when it involves a helpful inductive bias or prior belief. Regularized regression approaches, such as ridge regression, incorporate a penalty term that complements the fit term by providing a constraint on the solution, akin to how Occam's razor favors solutions that both fit the observed data and are simple. By incorporating such constraints or prior beliefs, the hope is that models will better predict future outcomes.

What makes a good prior belief or inductive bias? In the case of ridge regression, the norm of the regression coefficients is shrunk toward zero \cite{hoerl1970ridge,tikhonov1943stability} to control model complexity and reduce overfitting. However, in many domains, zero is not a reasonable \textit{a priori} guess for the true association between variables. For example, it would be strange to a priori predict the quality of a new home would be unaffected by the experience of the workers, quality of materials, reputation of the architect, etc. Because the world is somewhat predictable, a prior centered on the origin (Figure~\ref{fig:new_models}) is inappropriate.

If not zero, where does one turn for a useful prior? One answer is to look to human behavior. Humans use an assortment of clever strategies for learning and decision-making that perform well even in conditions of low knowledge. Simple heuristics that are fast and frugal \cite{Czerlinski1999} excel when training examples are scarce \cite{parpart2018heuristics}. People can also shift to more complex strategies when resources are available \cite{rieskamp2006ssl}. With increasing experience and expertise, humans often acquire a sophisticated understanding of domains.

Although heuristics are efficient and robust models in their own right, we propose they are a useful starting point or prior for more complete characterizations of domains. Advantages of heuristics include their ecological validity \cite{Czerlinski1999,Mata2012} and robustness across decision problems. Their weakness is insensitivity to aspects of the data due to their rigid inductive bias \cite{geman1992neural,parpart2018heuristics}. This weakness is ameliorated when heuristics function as priors within more complex models because priors can be overcome by additional data, much like how human experts develop more complex and nuanced knowledge with increasing experience in a domain. When data are abundant, the encompassing model would master the subtleties of the domain, whereas when data are scarce the heuristic prior would help guide predictions and increase robustness. Because the heuristics themselves are interpretable models, the solution of the encompassing model could be understood in terms of deviations from the heuristic prior.

\section{Robust priors based on decision-making heuristics}

\begin{figure}[htbp!]
\centering\includegraphics[width=0.8\linewidth]{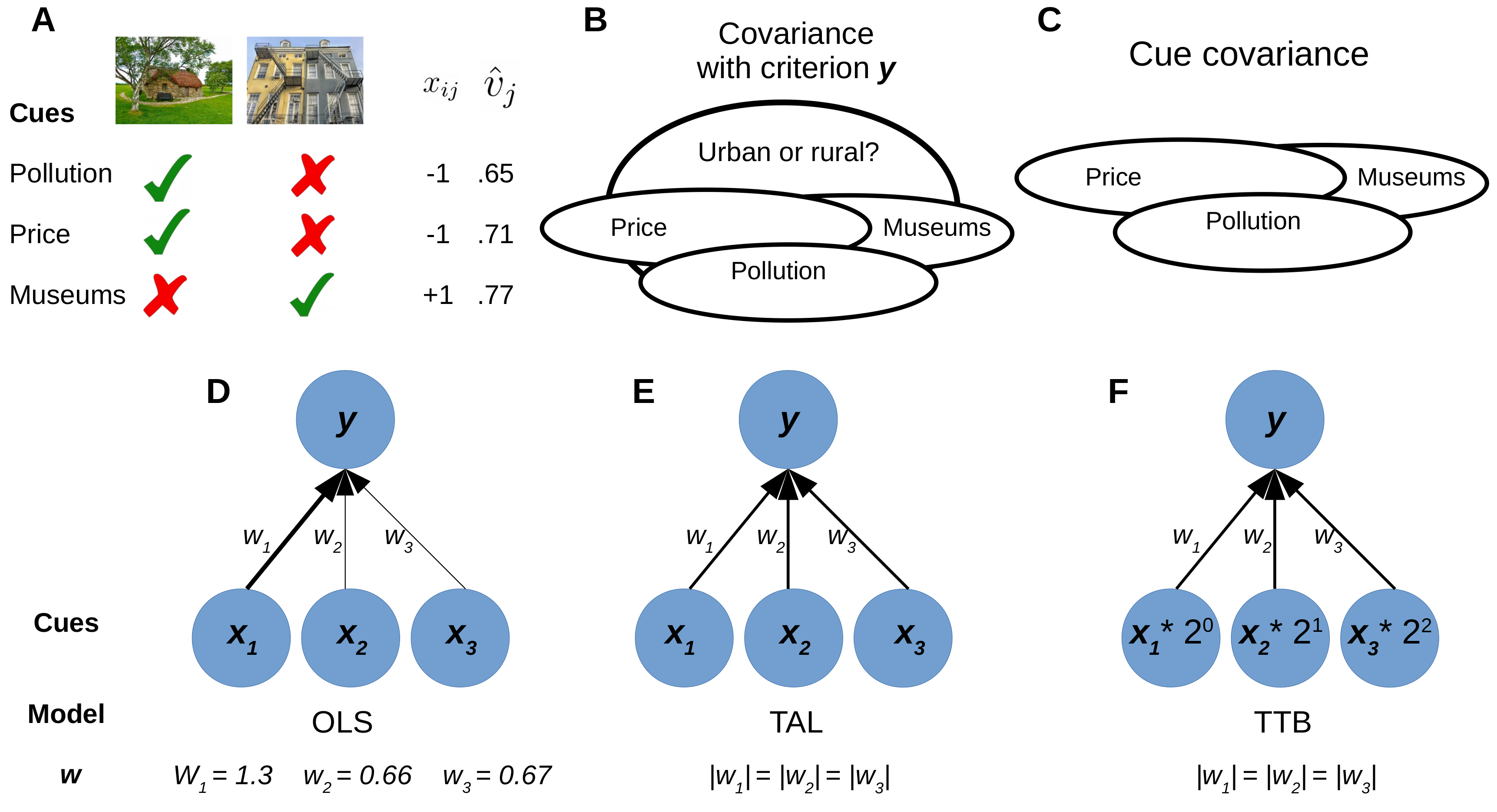}
\caption{TAL and TTB decision heuristics. (\textbf{A}) A hypothetical decision --- choosing between a rural ($-1$) or urban ($+1$) home based on cues ordered by cue validity: low pollution, low price, and proximity to museums. Each cue is coded as $-1$ when favoring the left option (rural), $+1$ for the right option (urban), and $0$ when the two options are equal on that cue. TAL sums cue values, choosing the rural home. TTB chooses based on the best cue (measured by $\hat{v}_j$, see Equation \ref{TAL2}) that distinguishes the options. Here, TTB would choose the urban home based solely on proximity to museums. (\textbf{B}) The covariance of the cues with the criterion $\bm{y}$ (urban or rural home), which $\hat{v}_j$ measures. (\textbf{C}) The covariance of the cues with one another; TAL and TTB heuristics disregard this information. (\textbf{D-F}) Illustrations of OLS, TAL and TTB. Here, OLS strikes a balance for correlated cues; low price and proximity to museums are (negatively) correlated. Thus, low pollution presents a higher weight $w_1$. TAL and TTB equate the absolute value of all weights. TTB additionally ranks and scales predictors according to their predictive value $\hat{v}_j$ in a non-compensatory way (multiplying cues by powers of 2).
}
\label{fig:heuristics}
\end{figure}

We used two well-known heuristics, tallying (TAL) and take-the-best (TTB) \cite{Czerlinski1999}, as priors in regularized regression models. These heuristics predict which of two options is preferable. For example, TAL or TTB could predict whether a rural or urban home is preferable based on several cues (Figure \ref{fig:heuristics}A). Each cue is ternary valued and indicates whether the left ($-1$) or right ($+1$) option is preferred on that dimension, with $0$ for a tie. TAL is a simple majority voting rule whereas TTB bases its decision on the single most predictive cue that can discriminate between the alternatives (both heuristics explained below; also, see Figure \ref{fig:heuristics}). 

\begin{figure}[htbp!]
\centering\includegraphics[width=1\linewidth]{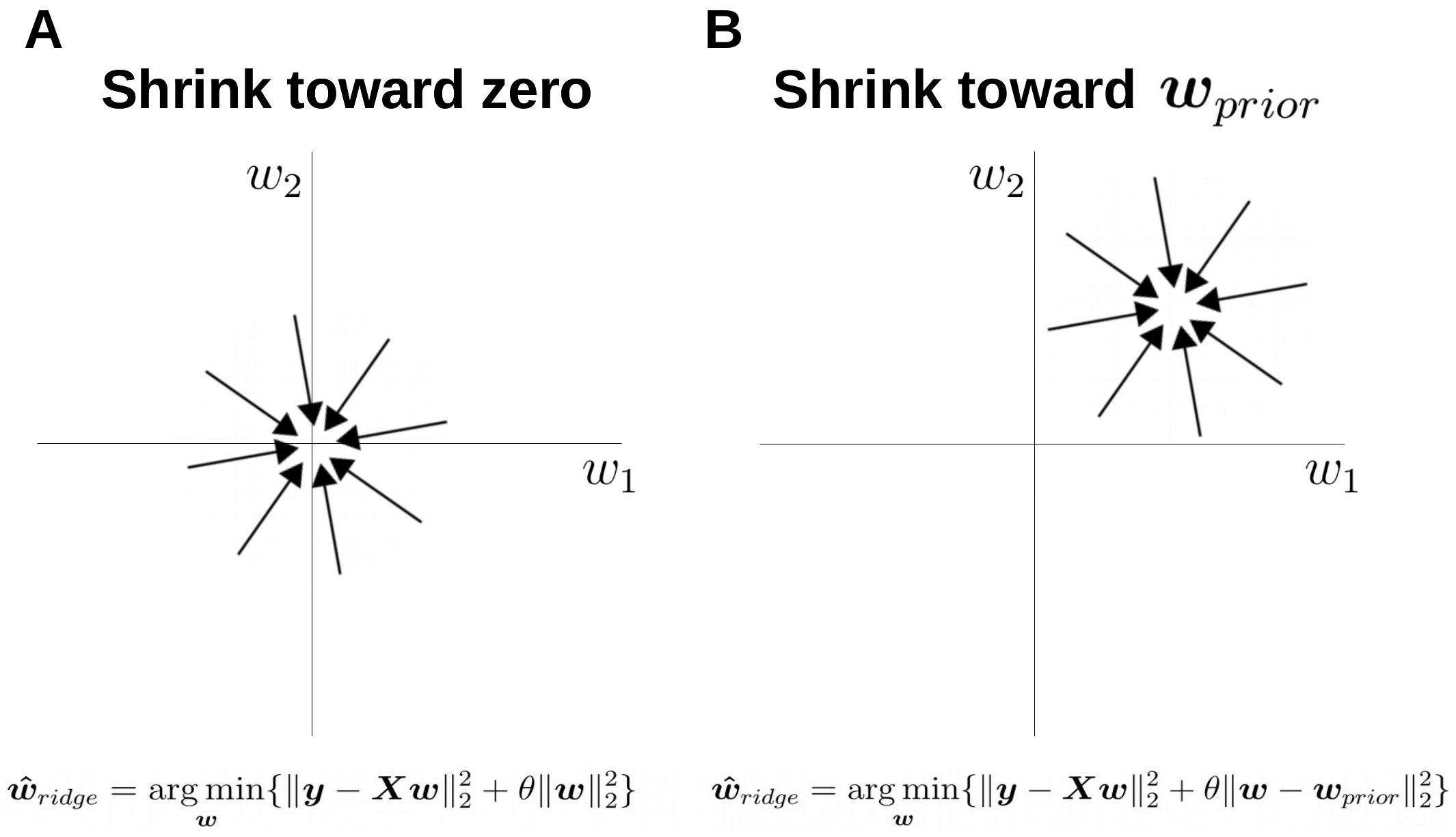}
\caption{Shrinking towards zero and non-zero priors. (\textbf{A}) Ridge regression shrinks weights towards $\protect\overrightarrow{\bm{0}}$ (Zero prior) in contrast to (\textbf{B}) a model using a prior based on the TAL heuristic where  all weights are equal and non-zero (i.e., $\bm{w}_{prior} \neq \protect\overrightarrow{\bm{0}}$). Using priors based on a heuristic (i.e., a constrained model) can increase robustness and interpretability. Equation \ref{eq:1} is shown at the bottom of panel \textbf{B}. The other equation simply drops $\bm{w}_{prior}$, equivalent to standard notation for ridge regression, here termed the Zero prior model.}
\label{fig:new_models}
\end{figure}

To use a heuristic as a prior, we use a two-step model-fitting procedure (cf. \cite{Zou2006TheProperties}). In step 1, we fit the heuristic to the training data. The resulting point estimate for the weight vector provides the penalty term (weighted by the penalty parameter $\theta$) within a regularized regression model in step 2. The penalty term shrinks regression coefficients toward the heuristic solution, as opposed to $\bm{\overrightarrow{0}}$ as in ridge regression (Figure~\ref{fig:new_models}). Increasing $\theta$ increases the strength of the prior, eventually pushing the regression solution to fully agree with the heuristic (cf. \cite{parpart2018heuristics}).

Our approach integrates heuristics with full-information (regression) models in a principled way that applies to a broad class of heuristics. The approach is to subtract a carefully constructed vector inside the penalty term of the well-known $L2$ cost function used in ridge regression. The cost function for standard ridge regression is 

\clearpage

\begin{equation}
\bm{\hat{w}}_{ridge} = \argmin_{\bm{w}} \{ \|\bm{y} - \bm{Xw}\|_2^2  + \theta\|\bm{w} - \bm{w}_{prior}\|_2^2\}, \label{eq:1}
\end{equation}

where $\| \cdot \|_2$ is the Euclidean ($L_2$) norm, $\bm{y}$ is the dependent variable $[y_1,...,y_n]^T$, $\bm{X}$ is an $n \times m$ matrix with one column for each of the $m$ predictor variables $\bm{x}_j$, $\bm{w}_{prior}$ is a column vector of zeros $\bm{\overrightarrow{0}} = [0_1,...,0_m]^T$, $\bm{w}$ is a vector of estimated regression coefficients $[w_1,...,w_m]^T$, and $\theta\ge0$ is a tunable penalty parameter.\footnote{Arrows over numerals and lowercase boldface are for vectors whereas uppercase boldface is for matrices or tensors. Carets are reserved for estimates, tildes for normalized scalars and overbars for arithmetic means.} Note that implicit priors are also found in regularized regression with all other norms ($L_n$) too, including LASSO regression ($L_1$). Thus, our insight generalizes to all other norms as well.

The first term inside the $\argmin$ of Equation \ref{eq:1} promotes goodness-of-fit in the model, whereas the second term --- known as the penalty term --- promotes smaller weights $\bm{w}$. As $\theta$ increases, the weights tend to $\bm{w}_{prior}$ in the limit (Figure \ref{fig:new_models}). (The derivation of the optimal weights $\bm{w}^*_{ridge}$ is included in the SI.) However, when $\theta = 0$, the model is equivalent to ordinary least squares (OLS) regression. OLS regression estimates coefficients without the penalty term:

\begin{equation}
\bm{\hat{w}}_{OLS} = \argmin_{\bm{w}} \{ \|\bm{y} - \bm{Xw}\|_2^2 \}\label{eq:ols}
\end{equation}

Normally, $\bm{w}_{prior}$ is not included in Equation \ref{eq:1}; it is implicit in more standard specifications of ridge regression, where the penalty term is written simply as $\theta\|\bm{w}\|_2^2$. Nevertheless, instead of a $\bm{\overrightarrow{0}}$ vector, one can generalize ridge regression with alternative constructions of $\bm{w}_{prior}$ (Figure \ref{fig:new_models}). 

As argued above, choosing this vector intelligently might improve learning of $\bm{\hat{w}}_{ridge}$ by imposing a more sensible inductive bias \cite{geman1992neural}. Although the decision-making literature has traditionally proposed that humans use certain classes of heuristics due to cognitive limitations \cite{bobadilla2018fast,Kahneman1974,Simon1978}, heuristics have also been justified from their ecological validity \cite{Czerlinski1999,dawes1979robust,parpart2018heuristics}. That is, the inductive biases they embody agree with the statistical structure of many natural environments, thus leading to better performance. Taking inspiration from the TAL and TTB heuristics, both in their success in describing human decision making \cite{bobadilla2018fast,Broder2003,Newell2003a,iris2018} and in application to real world statistical problems \cite{Czerlinski1999,parpart2018heuristics}, we propose a construction of $\bm{w}_{prior}$ based on these heuristics.

Below we discuss how to construct priors from TAL and TTB. We then report three applications. In the first application, we compared generalization performance (test set accuracy) and interpretability of model solutions on 20 classical datasets previously used in the decision-making literature \cite{Czerlinski1999,parpart2018heuristics}. There, the decision-making problem was to choose the better item within a pair (see Figure \ref{fig:heuristics}A and below). In the second application, we evaluated our approach within a classification paradigm in which a single item is assigned to one of two classes (e.g., friend or foe?). In the final application, we demonstrated the generality and benefits of our approach by analysing simulated brain imaging data where the prior is derived from a technique \cite{mumford2012deconvolving} that seeks to minimize collinearity amongst predictors in a manner that parallels how we derive heuristic priors.

\subsection{TAL and TTB Heuristics}

TAL and TTB do not adapt their form or complexity in light of the data.
For example, TAL is an equal-weights algorithm that uses only the signs of the coefficients \cite{Czerlinski1999,dawes1979robust}: the estimated weights $\hat{w}_j$ are constrained to be either $-1$ or $1$ (Figure \ref{fig:heuristics}E). 

The Tallying decision rule (TAL) is defined as

\begin{equation}
\hat{y}_i = \sign\left(\sum_{j}^{} \sign(\hat{v}_jx_{ij}) \right) \label{TAL1} \\
\end{equation}

\begin{equation}
\textrm{where } \; \hat{v}_j = \frac{R-W}{R+W} \in [-1,1] \label{TAL2}
\end{equation}

A cue's estimated cue validity $\hat{v}_j$ is defined as the difference between the numbers of correct predictions $R$ and incorrect predictions $W$, divided by the total number of predictions across all observations $R+W$ \cite{martignon1999does}. \footnote{Previous literature has defined cue validities as $R/(R+W)$. The definition used here is a linear transformation that simplifies description of the models.} Observations that do not present a prediction (i.e., $x_{ij} = 0$) are ignored. Notably, cue validities depend only on the relationship between each cue and the outcome, and not on covariance between cue. Thus, the definition of $\hat{v}_j$ is what makes heuristics insensitive to cue covariance. When assessing model performance, the validities $\hat{v}_j$ are estimated for each training set. 

The Take-the-Best (TTB) decision rule is

\begin{equation}
\hat{y}_i = \sign  (x_{ij^*})  \label{TTB1}
\end{equation}

\begin{equation}
\textrm{where } j^* = \argmax_{j} \left\{\mid \hat{v}_j \mid:x_{ij} \neq 0\right\} \label{TTB2}
\end{equation}
Whereas TAL sums the signs of the predictors to determine its response, TTB relies on the top predictor that differentiates the two options. When there is no evidence for either option, both TAL and TTB choose randomly (with probability $p = 0.5$ for each option). This occurs for TAL when Equation~\ref{TAL1} yields $0$ and for TTB when every $x_{ij}$ equals $0$.

We now define $\bm{w}_{prior}$ based on the TAL heuristic, referred to as $\bm{w}_{TAL\ prior}$. First, we determine a scalar coefficient
$\hat{w}_{TAL\ scale}$ shared across all predictor variables $\bm{x}_j$:

\begin{equation}
\hat{w}_{TAL\ scale} = \argmin_w \{ \|\bm{y} - \bm{X}(\bm{\hat{q}}w)\|_2^2\}.
\end{equation}

This equation is the same as Equation \ref{eq:ols} except that the vector $\bm{w}$ has been replaced by the product of a scalar $w$ and a column vector $\bm{\hat{q}}$, which has cue directionalities $\hat{q}_j=\sign(\hat{v}_j)$. Using this scalar, we define:
\begin{equation}
\bm{w}_{TAL\ prior} = \bm{\hat{q}}\hat{w}_{TAL\ scale}.
\end{equation}

To construct $\bm{w}_{TTB\ prior}$, we build from the intuition that TTB is equivalent to a noncompensatory weight vector such as $2^{\bm{\hat{r}}}$, where $\bm{\hat{r}}$ is a vector of ascending ranks for the absolute cue validities, $\hat{r}_j=|\{j':|\hat{v}_{j'}|<|\hat{v}_j|\}|$. Paralleling the definition of the $\bm{w}_{TAL\ prior}$, for the  $\bm{w}_{TTB\ prior}$, we also determine a shared scalar:

\begin{equation}
\hat{w}_{TTB\ scale} = \argmin_{\bm{w}} \{ \|\bm{y} - \bm{X}_{TTB}(\bm{\hat{q}}w)\|_2^2\} \label{ttb-scale}
\end{equation}

\begin{equation}
\bm{w}_{TTB\ prior} =\bm{\hat{q}}\hat{w}_{TTB\ scale}.
\end{equation}
However, we have a new design matrix $\bm{X}_{TTB}$ in Equation \ref{ttb-scale}, defined by

\begin{equation}
\bm{X}_{TTB} = \bm{X}\hat{\bm{P}}
\end{equation}

with $\hat{\bm{P}}$ being a diagonal matrix

\begin{equation}
\hat{\bm{P}} = diag(\phi^{\bm{\hat{r}}}), \quad \textrm{where } \phi := 2. \label{eq:ranks}
\end{equation}

This transformation has the effect of encoding cue validity directly into the design matrix by scaling each regressor according to a geometric progression.
In order for $\bm{w}_{TTB\ prior}$ to function appropriately as a $\bm{w}_{prior}$, the original design matrix  $\bm{X}$ is replaced with $\bm{X}_{TTB}$ in Equation \ref{eq:1}. When instead $\phi := 1$, we recover the TAL prior. Note also that this entire procedure is nearly equivalent to working with the original design matrix $\bm{X}$ and taking $\bm{w}_{TTB\ prior}$ to be proportional to $\hat{\bm{P}}\hat{\bm{q}}$ (the vector of signed exponentiated ranks, $\hat{q}_j \phi^{\hat{r}_j}$) rather than to $\hat{\bm{q}}$, except that the weights are differentially penalized according to their ranks. 

With these priors defined, we can now formally specify two regularized regression models. The TAL-prior model is defined by Equation \ref{eq:1} with $\bm{w}_{prior} = \bm{w}_{TAL\ prior}$. The TTB-prior model is defined by Equation \ref{eq:1} with $\bm{w}_{prior} = \bm{w}_{TTB\ prior}$ and with $\bm{X}$ replaced by $\bm{X}_{TTB}$.

In contrast to OLS, the use of the common scalar for all cues in the prior for both TAL-prior and TTB-prior highlights that both heuristics are insensitive to cue covariance information (see Figure~\ref{fig:heuristics}E,F). For TAL-prior, the common scalar reflects the fact that TAL is a (fully) compensatory strategy, whereas the design matrix in TTB-prior, $\bm{X}_{TTB}$, reflects the fact that TTB is a non-compensatory strategy. Later, we will evaluate how these differing priors affect the nature of penalized regression solutions.

\subsubsection{Logistic ridge regression}

The first two applications reported here use logistic ridge regression \cite{le1992ridge,schaefer1984ridge,van2015lecture}. To estimate weights for penalized logistic regression,  $\bm{\hat{w}}_{log(ridge)}$, we first obtain a scale parameter for an unpenalized logistic regression via maximum likelihood, where as above the weight vector is constrained to be proportional to the cue directionalities:
\begin{equation}
\hat{w}_{log\ scale} = \argmax_{w} \Pr[\bm{y}\vert\bm{X},\bm{w}=w\bm{\hat{q}}].
\end{equation}

The likelihood for logistic regression is as usual:
\begin{equation}
    \Pr[y_i\vert\bm{X}_{i\cdot},\bm{w}]=
    \begin{cases}
    \frac{1}{1+\exp(\bm{X}_{i\cdot}\bm{w})},&y_i=0\\
    \frac{1}{1+\exp(-\bm{X}_{i\cdot}\bm{w})},&y_i=1,
    \end{cases}
\end{equation}

where $\bm{X}_{i\cdot}$ denotes the $i^{\rm{th}}$ row of $\bm{X}$. We then define $\bm{w}_{log\ prior}$ as

\begin{equation}
\bm{w}_{log\ prior} = \bm{\hat{q}}\mid\hat{w}_{log\ scale}\mid.
\end{equation}

We then insert $\bm{w}_{log\ prior}$ into our final objective function for regularized logistic regression:
\begin{equation}
\bm{\hat{w}}_{log(ridge)} = \argmax_{\bm{{w}}} 
\left\{\Pr[\bm{y}\vert\bm{X},\bm{w}]
-\tfrac{1}{2}\theta\|\bm{w} - \bm{w}_{log\ prior}\|_2^2\right\}.
\label{eq:wlogridge}
\end{equation}

See Supplementary Information (SI) for an approximation of $\bm{\hat{w}}_{log(ridge)}$ for regularized logistic regression via the Newton-Raphson iterative algorithm.

\section{Application I: Heuristic decision making}
\label{S:3}

Regularized regression models with heuristic priors were evaluated on the 20 datasets that have been previously used to compare heuristics with ordinary least squares (OLS) regression \cite{Czerlinski1999,Katsikopoulos2010,parpart2018heuristics}. For each of the 20 problems, the cues for the two options on each trial were binary valued (see Methods below for more details), which leads to ternary-valued inputs according to our coding scheme (see Figure~\ref{fig:heuristics}A). 

\subsection{Methods}

The preprocessed data were retrieved from an Open Science Foundation (OSF) repository \cite{ds20}, used to evaluate the half-ridge and COR models by \cite{parpart2018heuristics}. 
In accord with previous research, cue attributes were dichotomized by median split \cite{Czerlinski1999,parpart2018heuristics}. 

The data were transformed into a format appropriate for decision-making problems where all pairwise comparisons between observations were encoded as the signed differences in (binary) attributes (possible values: -1, 0, and +1). The decision in our coding scheme is $-1$ for the left choice and $+1$ for the right choice, which was mapped to $0$ and $1$ for logistic regression: e.g., in the homestead example (Figure \ref{fig:heuristics}A), rural is coded as -1 (or 0 for logistic models) and urban is coded as +1.
This is common procedure in the decision-making literature \cite{Czerlinski1999,Katsikopoulos2010,parpart2018heuristics}. Formally, this consists of training pairs $(\bm{x}_1,y_1),...,(\bm{x}_n,y_n)$ with $\bm{x}_i \in \{-1,0,1\}^m$. Training sets consisted of 50 training pairs from which the priors were learned from. All results were computed for 1000 iterations (i.e., different partitions into training and test sets) for all penalty values. 


As the penalty parameter $\theta$ increases, the penalized regression models with the $\bm{w}_{TAL\ prior}$ and $\bm{w}_{TTB\ prior}$ converge to their corresponding heuristics (Figure \ref{fig:app1_results}A). As a sanity check, the TAL-prior and TTB-prior models were validated on simulated data in the SI (Figures S1 and S2, respectively) by tracking their agreement with OLS predictions. Effectively, agreement with OLS is higher for low penalty values and agreement with TAL-prior or TTB-prior is higher for high penalty values. Individual plots for each of the twenty datasets are also included in the SI (Figures S3 and S4).

Although regression models are interpretable in that each feature's importance follows from its weight, the heuristic penalty terms make clear how the prior shapes the solution and how the solution differs from the prior, which itself is an interpretable solution. To evaluate how the form of the solution changes as a function of the prior, we calculated normalized Shannon entropy defined as:

\begin{equation}
    \tilde{H} = \frac{-\sum_j{\tilde{w_j} log_2 \tilde{w_j} }}{ -\sum_j{
    \frac{1}{m}  log_2 \frac{1}{m}  }} \label{entropy.eq}
\end{equation}
where $\tilde{w_j}$ is 

\begin{equation}
    \tilde{w_j} =\frac{|\hat{w}_j|}{\|\bm{\hat{w}}\|_1}\phi^{\hat{r}_j},
\end{equation}

$\phi := 1$ for TAL-prior and $\phi := 2$ for TTB-prior, and $\| \cdot \|_1$ is the $L_1$ norm, such that $\tilde{H} \in [0,1]$ for any number ($m$) of predictors. Equation~\ref{entropy.eq} provides an intuitive measure of how compensatory a solution is. The measure will peak at $1$ when the predictive force of the weights is uniform, as in the TAL heuristic. 

\subsection{Results}

\begin{figure}[htbp!]
\centering\includegraphics[width=0.7\linewidth]{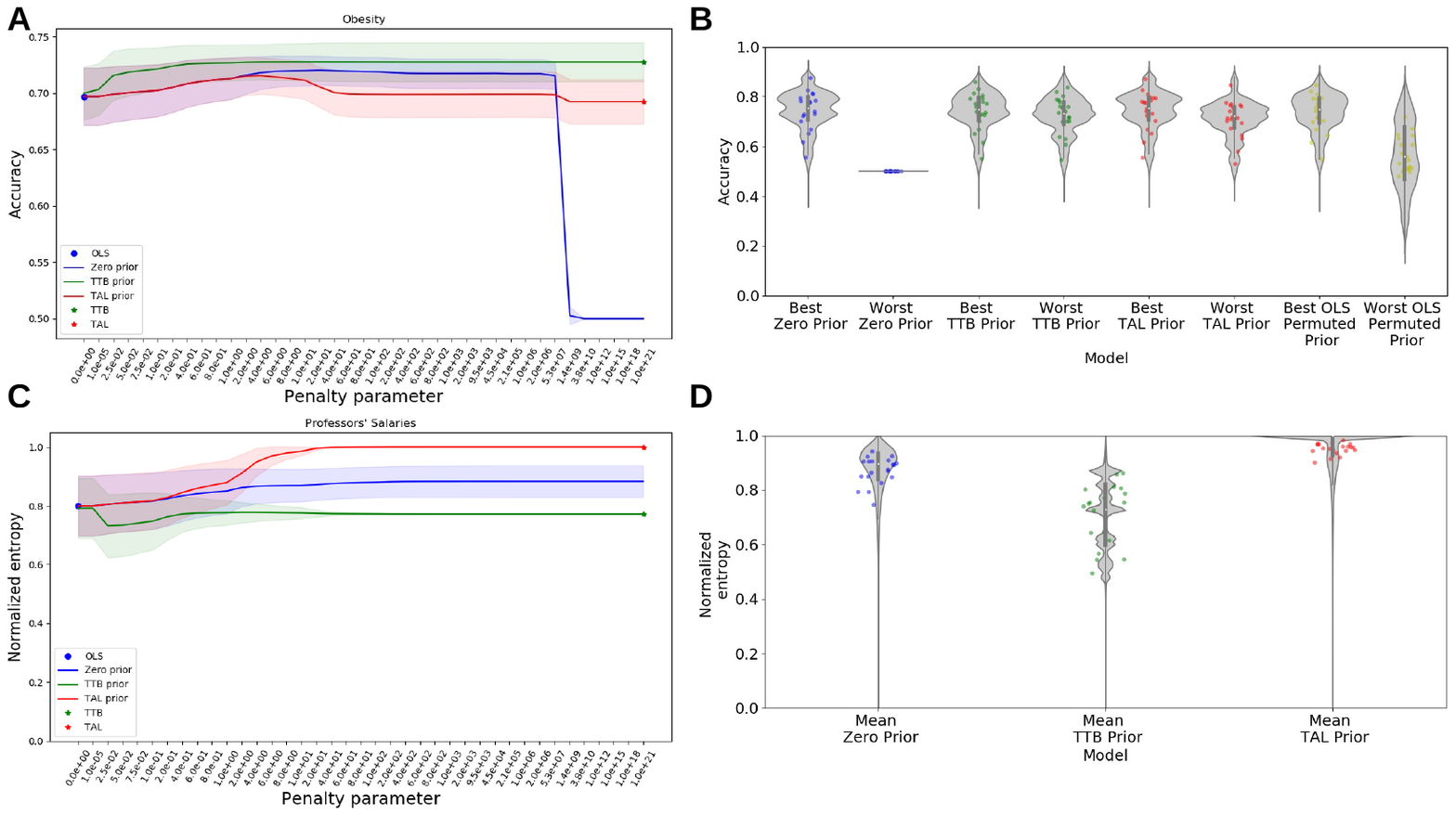}
\caption{Accuracy and normalized entropy for the 20 datasets in Application 1. Training set size was fixed at 50. (\textbf{A}) Test set accuracy across penalty values for the Obesity dataset. At low penalties, the models all agree with OLS (unpenalized). Under strong penalties, the Zero prior model (standard ridge regression) converges to chance performance as weights shrink toward $\protect\overrightarrow{\bm{0}}$. TAL-prior and TTB-prior models converge toward their respective heuristics and are robust. (\textbf{B}) Test set accuracy for all 20 datasets for best and worst performing penalty values for each model (see SI). The OLS permuted prior model is a penalized regression model with a permuted OLS solution as prior. Heuristic prior models are most robust. (\textbf{C}) Normalized entropy (Equation~\ref{entropy.eq}) for the Professors' Salaries dataset, which measures how compensatory the weights are. The TAL-prior model becomes maximally compensatory as penalty increases, unlike the TTB-prior model.  (\textbf{D}) Normalized entropy for all 20 datasets across all penalty values, which orders as TAL-prior, Zero prior, TTB-prior. For \textbf{B} and \textbf{D}, violins represent density estimates for the respective metric. Each dot is one of 20 datasets. For \textbf{A} and \textbf{C}, shaded areas represent 1 standard deviation.} 
\label{fig:app1_results}
\end{figure}

As predicted, these models are robust across the range of $\theta$ values because they converge to a reasonable estimate (i.e., a sensible heuristic). In contrast, while ridge regression performs well overall, its performance suffers at higher penalty values as its weights are pulled toward $\bm{\overrightarrow{0}}$. The robustness of the penalized regression models with heuristic priors held across the 20 datasets (Figure \ref{fig:app1_results}B). Notice that regularization using any nonzero prior is not sufficient for robustness -- an ad hoc nonzero prior (OLS Permuted Prior) was not robust. The OLS permuted prior model is a penalized regression model with a permuted OLS solution as prior (i.e., where the weights from the OLS solution have been permuted).

We confirmed that $\tilde{H}$ for the TAL-prior model would converge to $1$ with increasing penalty $\theta$, in contrast to the TTB-prior model. We also predicted ridge regression's $\tilde{H}$ would be somewhat lower than TAL-prior's. That is, convergence to a $\bm{\overrightarrow{0}}$ weights vector for standard ridge regression is nonsensical and effectively resisted in the optimization, providing more heterogeneous weights than otherwise expected. These predictions held (Figures~\ref{fig:app1_results}C,D). 

The results presented here hold under an alternative training scheme where we also evaluate OLS as a prior itself (\textit{Splitting training data} in the SI). OLS performs worse as a prior on the majority of datasets (Figure S7 in the SI) and also shows higher variance overall (Figure S8 in the SI).

In these 20 decision problems, models using priors based on TAL and TTB were robust across the entire range of prior strengths. These penalized regression models shrunk to a reasonable prior based on a simple heuristic that discards covariance information amongst predictive cues. The forms of the solutions were interpretable and followed from the priors.

\section{Application II: Breast Cancer Classification}
\label{S:4}

In this application, we conducted the same analyses as in Application I, but for a classification problem as opposed to a forced choice between two options. We applied the models to the Breast Cancer Wisconsin (Diagnostic) Data Set from the UCI data repository \cite{uci_repo}. In this task, models predicted whether an item was cancerous or not based on binary features (see Methods for more details). The predictors were discrete as in Application I, though the identical approach would apply to continuous predictors or to a mixture of discrete and continuous predictors.\footnote{Our approach easily generalizes to the continuous setting by placing predictors on the same scale through some form of normalization (e.g., \textit{z}-scoring).}

\subsection{Methods}

The data comprises nine cues ($m = 9$) that describe characteristics of the cell nuclei present in digitized images of fine needle aspirate (FNA) of breast masses \cite{uci_repo}. Data points with missing cue values were removed, resulting in a total of $n = 478$ observations. All variables were binarized by median split.

\begin{figure}[htbp!]
\centering\includegraphics[width=0.7\linewidth]{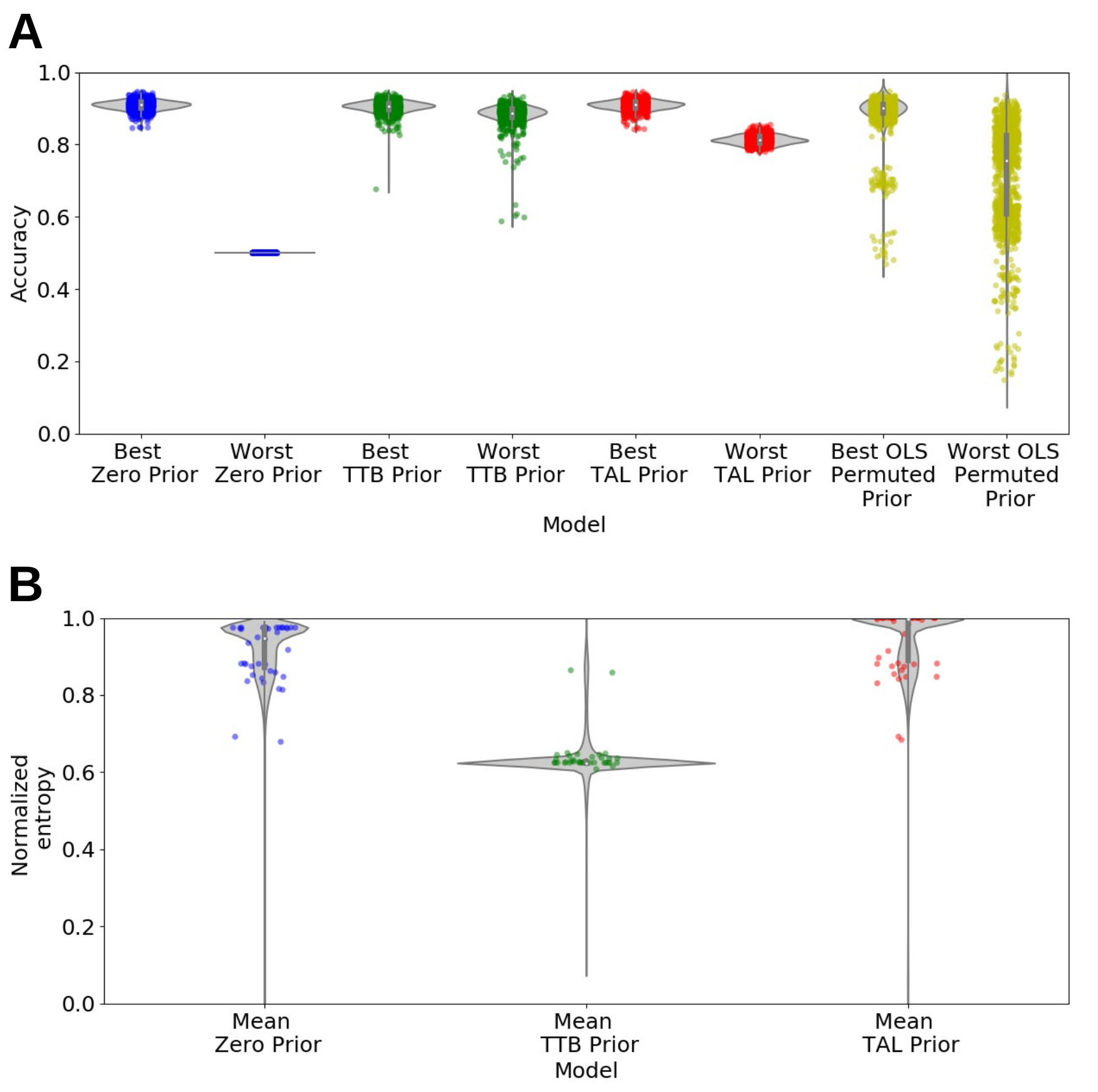}
\caption{Generalization performance and normalized entropy for Breast Cancer Wisconsin (Diagnostic) Data Set after training on 100 items (Application II). The same models are considered here as in Application I (see Figure~\ref{fig:app1_results}). (\textbf{A}) Test set accuracy for the best and worst performing penalty value for each model (see SI). The key finding is that models with heuristic priors are most robust.
(\textbf{B}) Normalized entropy (Equation~\ref{entropy.eq}) averaged across the range of penalty values reflects how compensatory a model's predictions are, led by TAL-prior, followed by the Zero prior, and finally the TTB-prior model. Each dot represents one of the tested penalty values averaged over 1000 train-test splits. The gray violins represent the respective density estimates in both panels.} 
\label{fig:bcancer100}
\end{figure}

In an analogous fashion to how we constructed $\bm{X}$ in Application I, here we transformed the original data by median splits. For each cue, if the value was equal to the median, it received a value of $0$, if it was above the median it was equal to $+1$ and if it was below the median it was equal to $-1$. Formally, this also produces training pairs $(\bm{x}_1,y_1),...,(\bm{x}_n,y_n)$ with $\bm{x}_i \in \{-1,0,1\}^m$. Training sets consisted of 100 training pairs from which the priors were learned from. However, we did not construct a matrix of pairwise comparisons of observations as before. The dependent variable was binary, $\bm{y} \in \{-1,+1\}$, coding for malignant and benign tumours, respectively. This preprocessing of the data $\bm{X}$ is closer to the way regression models are calculated for everyday applications. 
Both the mean test accuracy (Figure \ref{fig:bcancer100}A and Figure S5 in the SI)
and the mean normalized entropy (Figure \ref{fig:bcancer100}B and Figure S6 in the SI)
were averaged over 1000 iterations for each penalty value.

\subsection{Results}

The results were in accord with Application I. The models with a heuristic prior were robust across the range of $\theta$ values (Figure \ref{fig:bcancer100}A). As in Application I, the priors shaped the form of the solution in the predicted manner (Figure \ref{fig:bcancer100}B) with the TAL-prior model having the most compensatory solutions.

\section{Application III: Estimation in Brain Imaging Analyses}
\label{S:5}

In Applications I and II, the task was to generalize from training items to make decisions about test items. In Application III, the objective was to estimate the weights themselves. We considered simulated functional magnetic resonance imaging (fMRI) time series that allowed for comparing estimates to ground truth.

Brain imaging datasets are challenging to analyze because they measure the brain's hemodynamic response, which is a temporal and spatially autocorrelated, high-dimensional, noisy, and time-lagged signal. The signal is composed of thousands of voxels (voluminous pixels) with coordinates in space ($\mathsf{x, y, z}$) and time ($n$ time-points). Correlations across space and time due to psychological (e.g., \cite{Visscher2009}), neurovascular \cite{Boynton2012} and physical \cite{smith1999investigation} effects complicate the independence and linearity assumptions used to model the signal in each voxel. Furthermore, the observed blood-oxygen-level dependent (BOLD) signal is only indirectly related to the outcome variable of interest (neural activity), via the hemodynamic response function (HRF), which is normally modelled as a double gamma function. 

In task fMRI, the BOLD time series for a voxel is modelled by weighting events, such as a sequence of pictures (e.g., dog, truck, face, etc.) presented to a study participant. In addition to nuisance regressors, one typically estimates a beta weight for each event (convolved with the HRF). We refer to this standard method as least squares all (LSA), which is unpenalized and plays a role analogous to OLS in Applications I and II.

However, for the reasons discussed above, collinearity in the time series can compromise parameter estimation \cite{mumford2015orthogonalization}, particularly in rapid event designs (e.g., trial duration of one or two seconds). One proposed solution, which we refer to as least squares separate (LSS), is to estimate a separate model for each event rather than a single model for all events \cite{rissman2004measuring}. Each model estimates one beta weight for the target event (i.e., trial) and a second shared beta weight for all other events \cite{turner2010comparison}. In practice, LSS produces better (less variable) estimates by being less sensitive to collinearity in the time series \cite{mumford2012deconvolving}.

We view LSS as analogous to the heuristics considered in Applications I and II. The TAL and TTB heuristics are insensitive to cue covariance. Specifically, cue validity and cue direction are estimated individually for each predictor. Moreover, we implemented these heuristics in a regression framework with a single beta weight (e.g., $\hat{w}_{TAL\ scale}$) to derive a prior. In both models, simplification is achieved by forcing multiple predictors to share a single regression weight. Analogously, each LSS model forces all but the target event (out of potentially hundreds of events) to share a common beta weight.

Like TAL and TTB, we predicted that LSS would provide an effective prior for a penalized regression model because it provides a reasonable and robust starting point to move from when the data warrant. We predicted that a penalized regression model with an LSS prior would outperform both LSS (high $\theta$) and the LSA approach ($\theta = 0$).

\subsection{Methods}

To build a continuum of models between LSA and LSS, we include the weights derived from LSS as a target (i.e., prior) in the penalty term within a regularized LSA model. Thus, the weights from the LSS-prior model are estimated with the following objective:

\begin{equation}
\argmin_{\bm{w}} \{ \|\bm{y} - \bm{X}_{LSA}\bm{w}\|_2^2  + \theta\|\bm{w} - \bm{w}_{LSS\ prior}\|_2^2\} \label{lssPrior}
\end{equation}

Paralleling our treatment of the decision heuristics as priors, Equation~\ref{lssPrior} specifies a continuum of models ranging from LSA ($\theta = 0$) to LSS ($\theta \rightarrow \infty$). For all models, $\bm{y}$ is the activation time series for a single voxel; with spatial indices (i.e., coordinates in brain space) its notation is $\bm{y}_{\mathsf{xyz}}$. 



Both LSA and LSS are known as massive univariate GLMs, since they model each voxel independently. For a given voxel, LSA estimates weights as

\begin{equation}
\bm{\hat{w}}_{LSA} = (\bm{X}_{LSA}^T\bm{X}_{LSA})^{-1}\bm{X}_{LSA}^T\bm{y}
\end{equation}

where $\bm{y}$ is the BOLD response time series for a voxel and $\bm{X}_{LSA}$ is the $n \times m$ design matrix with number of columns $m$ equal to the number of trials $\ell$, with only one event per trial. (Each $\bm{x}_j$ is an event for LSA, but this changes for LSS.) 
This means that a column in $\bm{X}_{LSA}$ models a single event in the experiment. The number of brain scans or time-points $n$ is usually larger than the number of trials (events) ($n > \ell$) because more than one brain scan is acquired per trial. Quite commonly, a regressor models an event (such as stimulus presentation) with a boxcar function, that models the duration of the stimulus, convolved with a double gamma HRF \cite{Boynton2012}. We will not focus here on how the regressors that model the BOLD signal are constructed. Instead, we focus on the GLMs that receive those regressors as input.


The LSS model differs from the LSA model in that the $\bm{X}_{LSA}$ matrix is replaced with a set of matrices $\bm{X}_{LSS_1},...,\bm{X}_{LSS_\ell}$ which results in one GLM per trial: 

\[
\hat{w}_{LSS_1} = \bm{s}[(\bm{X}_{LSS_1}^T\bm{X}_{LSS_1})^{-1}\bm{X}_{LSS_1}^T\bm{y}]
\]

\begin{equation}
\vdots
\end{equation}

\[
\hat{w}_{LSS_\ell} = \bm{s}[(\bm{X}_{LSS_\ell}^T\bm{X}_{LSS_\ell})^{-1}\bm{X}_{LSS_\ell}^T\bm{y}]
\]

Each $\bm{X}_{LSS_k}$ has dimensions $n \times 2$, where $n$ is the same as before. Each weight $\hat{w}_{LSS_k}$ is selected as the first coefficient from its respective GLM, via mutipication by $\bm{s} = [1 \quad 0]$. Each  $\bm{X}_{LSS_k}$ is constructed as mentioned above, with the first predictor variable $\bm{X}_{LSS_k}\bm{s}^T$ modelling a single experimental trial of interest (i.e., the $k$th trial) and the second predictor being a nuisance variable $\bm{X}_{LSS_k}[0 \quad 1]^T$ 
modelling all other trials in the experiment (i.e., all $\ell - 1$ trials excluding trial $k$). 
The LSS-prior model in the main text uses $\bm{\hat{w}}_{LSS}$ as $\bm{\hat{w}}_{LSS\ prior}$.

\subsubsection{Simulated fMRI Data}

There were 1000 simulations performed for each of 9 different designs (see below) with varying levels of signal-to-noise ratio (SNR) and interstimulus intervals (ISI; time between events). 
The simulations were performed on modified code from the rsatoolbox \cite{Nili2014}, which can be consulted at:

https://github.com/bobaseb/rsa\_toolbox\_lss/tree/develop/LSS\_project 

Each simulation consisted of a cluster (i.e., region of interest) of task-sensitive signal voxels with observed data generated for all $\ell$ trials by weights $\bm{\Psi}\in{\rm I\!R}^{m \times d \times d \times d}$, where $m = \ell$ and each spatial dimension $d = 7$. The weights $\bm{\Psi}$ were embedded in an array tripled along each spatial dimension $\bm{\Omega}\in{\rm I\!R}^{m \times 3d \times 3d \times 3d}$ (i.e., the simulated brain). The weights for non-task-sensitive voxels in $\bm{\Omega}$ (i.e., those not in $\bm{\Psi}$) were set to zero. 
Scanner noise $\bm{E} \in {\rm I\!R}^{n \times 3d \times 3d \times 3d}$ had entries $\epsilon{_{i\mathsf{xyz}}}$ drawn i.i.d. from a centered normal distribution $\mathcal{N}(0,\sigma^2_{scanner})$, where $\sigma^2_{scanner} = 10000$, and was added to $\bm{X}_{LSA}\bm{\Omega}$ to generate the observed signal: 

\begin{equation}
\bm{Y} = \bm{X}_{LSA}\bm{\Omega} + \bm{E}. \label{fMRI_sim}
\end{equation}
Thus, for a single voxel described by a set of spatial coordinates $\mathsf{x, y, z}$, we have data across time in $\bm{Y} \in {\rm I\!R}^{n \times 3d \times 3d \times 3d}$, represented as $\bm{y}_{\mathsf{xyz}}$. For observations $\bm{Y}$, the subset corresponding to voxels that are task-sensitive is denoted as $\bm{Y}_{\bm{\Psi}}$. Notice the use of $\bm{X}_{LSA}$ to generate simulated data instead of $\bm{X}_{LSS}$. In fact, there is no straightforward way to construct weights $\bm{\Psi}$ (embedded in $\bm{\Omega}$) to multiply with the set of matrices $\bm{X}_{LSS_k}$. 

To simulate spatiotemporal correlations in the data, the scanner noise $\bm{E}$ was smoothed along its four axes for each run (two runs total, see below), using a Gaussian spatiotemporal smoothing kernel with full width at half maximum (FWHM) equal to 4 mm for the three spatial dimensions and 4.5 s for the temporal dimension. (Voxel size was set in millimeters at the default value of $3\times3\times3.75$ in the rsatoolbox.)

For each simulation, each coordinate of the effect center ($\mathsf{x, y, z}$, as defined in the rsatoolbox) --- where the signal voxels were placed inside the simulated brain $\bm{\Omega}$ --- was uniformly sampled between 1 and 11 inclusive.
Two separate runs ($r = 2$) were simulated on each of the 1000 iterations and each run had 20 repetitions of each of two stimulus types ($\ell = 20s \textrm{ and } \mathsf{s} = 2$).
Simulating more than one run and stimulus type contributes to the ecological validity of the simulation, especially for studies that focus on classification (MVPA) where one run is used for training and another for testing. Repetition time (TR; duration for obtaining one full brain scan) was set to 1 second and event duration (ED, the duration of a stimulus on the screen in the 
MRI scanner) was set to 1.5 seconds. A trial's duration is given by $ED + ISI$. There are also $\lceil\ell/3\rceil$ null epochs, randomly interspersed with the trials, where no stimulus is shown, each with a duration of $ED + ISI$ seconds. This kind of experimental design is common because it further helps reduce collinearity between trials and aid in the estimation of $\bm{w}$. Thus, for each run $n \gtrapprox \frac{\frac{4}{3}\ell \times (ED+ISI) + \mathsf{t}_{end}}{TR} $, where $\mathsf{t}_{end}$ is a temporal slack after the last trial that allows the BOLD signal enough time to decay. The exact number of time-points $n$ depends on the HRF model that was used \cite{Boynton2012}. This information is encoded in the design matrix $\bm{X}_{LSA}$.

\begin{figure}[htbp!]
\centering\includegraphics[width=1\linewidth]{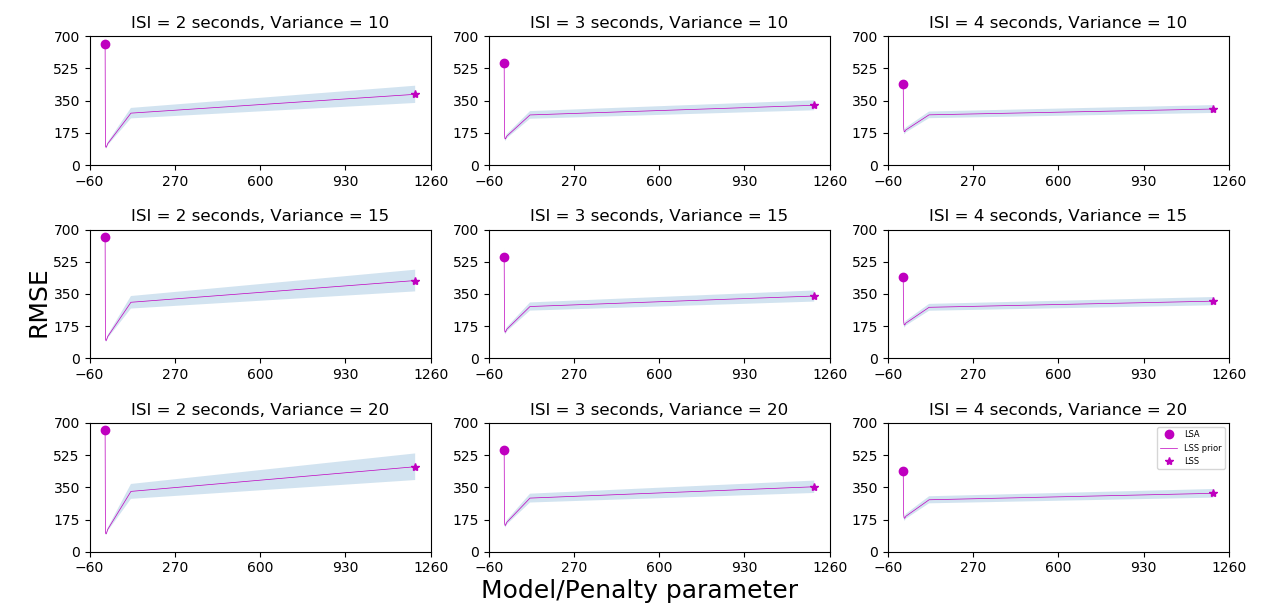}
\caption{The fMRI simulation results. Root mean squared error averaged over voxels and simulations ($\overline{\rm RMSE}$) between estimated regression weights $\bm{\hat{w}}$ of each model and the true data-generating weights $\bm{\Psi}$ (see Methods) for different interstimulus intervals (ISI) and variances (signal-to-noise ratio, SNR: $\sigma^2_{\bm{\Psi}}$, see Methods). The 3x3 design is presented such that each row displays a different level of variance while each column displays a different ISI. Our LSS-prior model is equivalent to the LSA model (purple circle) when $\theta=0$ and the LSS model (purple star) when the $\theta$ penalty parameter is large. Shaded areas represent three standard deviations above and below the estimate. Lower $\overline{\rm RMSE}$ values convey superior performance. The key finding is our penalized regression model is superior to the standard LSA model and the LSS model (which serves as the prior) for moderate prior strength $\theta$. For all panels, the results for all penalty values (in each model and dataset) are averaged over 1000 train-test splits.}
\label{fig:fmri_sims}
\end{figure}

To sample the data-generating weights $\bm{\Psi}$ with correlations between (task-sensitive) voxels, we did the following: For each of the $\ell/\mathsf{s} = 20$ trials per stimulus, we sampled from $\mathcal{N}(\bm{\mu},\bm{\Sigma})$. 
Each entry $\mu_1,...,\mu_{d^3}$ in mean vectors $\bm{\mu}_1$ and $\bm{\mu}_2$ (for each stimulus, respectively) was i.i.d., drawn from a normal distribution $\mathcal{N}(0,\sigma^2_{\bm{\Psi}})$ for three levels of SNR ($\sigma^2_{\bm{\Psi}} \in \{10,15,20\}$). These were sampled for each iteration (a thousand iterations total) in each of the nine designs (Figure \ref{fig:fmri_sims}) but kept constant across runs. 
The covariance matrix $\bm{\Sigma}$, with dimensions $d^3 \times d^3$, induces the correlations between task-sensitive voxels and was kept constant across runs but resampled on different iterations. 
It was drawn from a scaled Wishart distribution $W(\bm{V},df)/df$ with degrees of freedom $df = d^3$. 
The symmetric positive definite matrix $\bm{V}$ was constructed with ones on the diagonal and $0.7$ for all off-diagonal values, representing a high degree of correlation between task-sensitive voxels. As presented in Figure \ref{fig:fmri_sims}, the 3x3 design of the simulations had three levels of ISI $\in \{2, 3, 4\}$ (in seconds) and three levels of SNR (as mentioned above). 

After sampling all $\ell \times d^3$ weights for a run, we have the object $\bm{M} \in {\rm I\!R}^{\ell \times d^3} $. 
This matrix of weights (trials by voxels) $\bm{M}$ was permuted along the temporal dimension $\ell$ and was arbitrarily mapped to the spatial coordinates of $\bm{\Psi}$ --- and by implication, of $\bm{\Omega}$ too--- such that $\bm{M} \to \bm{\Psi} \to \bm{\Omega}$, before applying Equation \ref{fMRI_sim}. 

\subsubsection{Model scoring}

Our evaluations of the models were done with the root mean squared error (RMSE) of each $\bm{\hat{w}}$ for each model (i.e., LSA, LSS, LSA-prior and LSS-prior models) with respect to the ground truth of each vector $\bm{\psi}_{\mathsf{xyz}}$ in $\bm{\Psi}$:

\begin{equation}
RMSE(\bm{\hat{w}}_{\mathsf{xyz}},\bm{\psi}_{\mathsf{xyz}}) = \sqrt{\sum_{j=1}^{\ell} \frac{(\hat{w}_j-\psi_j)^2}{\ell}}
\end{equation}

averaged across all the weights for task-sensitive voxels in $\bm{\Psi}$ and the 1000 iterations of simulated data:

\begin{equation}
\overline{\rm RMSE} = \sum_{1}^{1000}\sum_{\mathsf{xyz}}^{} \frac{RMSE(\bm{\hat{w}}_{\mathsf{xyz}},\bm{\psi}_{\mathsf{xyz}})}{1000d^3}
\end{equation}

\subsection{Results}

Our main prediction held (see Figure~\ref{fig:fmri_sims}). Across a range of task conditions, our penalized regression approach with $\bm{w}_{LSS\ prior}$ outperformed both LSS (equivalent to large $\theta$) and LSA (equivalent to $\theta =0$) for intermediate penalty values of $\theta$ (Figure \ref{fig:fmri_sims}). LSS provided an effective prior for our penalized regression. Replicating previous work, $\overline{\rm RMSE}$ was lower for LSS than LSA, akin to less-is-more effects in which decision heuristics can best OLS (e.g., TTB in Figure~\ref{fig:app1_results}A).

\section{General Discussion}
 
We looked toward human decision making to identify an effective prior for regularized regression and found that decision heuristics which disregard cue covariance information offer a number of advantages, such as robustness and interpretability. These heuristics offered a sensible starting point compared to the usual way of defining $\bm{w}_{prior}$ for most ridge regression applications (i.e., as the zero vector).

Here we have presented three different types of applications in over twenty different datasets, germane to the fields of decision making, fMRI analysis, and statistical modelling. We have validated the utility of using heuristics like TAL and TTB to construct $\bm{w}_{prior}$, as well as using other algorithms which lack a normative foundation and parallel the operation of heuristics --- like LSS in the case of fMRI time series modelling.

Three main benefits of no-covariance priors are worth highlighting. First, predictions using a no-covariance prior are likely to provide at least as good, if not better, predictions than a vector of zeros as coefficients. Examples of the TTB-prior model outperforming other models are seen in Figure \ref{fig:app1_results}A and in the lower $\overline{\rm RMSE}$ values obtained in Application III. 

Second, catastrophic failure of the model is avoided for extremely high values of $\theta$, whereas in normal ridge regression, convergence to the zero vector for high penalty values results in essentially random guessing for the comparison and classification tasks presented here. Convergence toward very small weights may also create implementation issues on digital computers which have limited precision. For example, differences in how floating point numbers are represented in supporting software libraries could reduce the reproducibility of results.

Third, this class of priors has theoretical significance. On the one hand, the model class introduced here further integrates the notions of heuristic decision making and full information algorithms along one continuum of models (as in \cite{parpart2018heuristics}). Choosing heuristic priors that contrast compensatoriness of the environment, like TAL and TTB do, helps us interpret both the solutions of our models and the environment itself in an easier way than is possible with OLS or the Zero prior model. Likewise, the solution of the encompassing model could be understood in terms of deviations from the heuristic prior. Other informative comparisons could be made to the OLS solution, including how it diverges from the heuristic prior. Finally, our framework provides a way to simulate fMRI data with LSS weights, previously not possible due to the arbitrariness of defining weights for the LSS nuisance variables (see Application III).

The theoretical contribution of this model class is worth emphasizing since it also provides a lens on why heuristics are useful in the first place. The priors offered by heuristics confer robustness; unlike the Zero prior, they embody a sensible inductive bias. This dovetails with why heuristics can operate defeasibly. Speculatively, humans and other cognitive agents may have evolved to implement these priors as a rule. Like Occam's razor, humans also show bias towards simple solutions for many decision-making tasks \cite{gigerenzer1999simple, Kahneman1974}. With expertise (i.e., acquiring more data), the solutions can change \cite{hornsby2014improved}, but initially, very general strategies like assuming independence among covariates have been documented \cite{broder2000assessing,Gigerenzer1996}. Of course, this is only one notion of expertise. Other notions could include less effort during inference or rule application, finding appropriate features of a domain and ease of searching for new strategies or creating new ones. Furthermore, experts are not even guaranteed to perform better than statistical techniques (cf. \cite{Meehl1954ClinicalEvidence.}).

Instead of being all-or-none, heuristic use may move along a continuum \cite{Newell2005} as a function of prior strength and experience. Indeed, heuristic use in human decision making is not without its caveats \cite{Newell2003}, as is their supposed frugality \cite{bobadilla2018fast,Dougherty2008}. What is clear is that no heuristic will be best in all environments (cf. no free lunch theorem). Instead, each heuristic is best suited to certain environments and can be seen as embodying a prior that reflects beliefs about the environment. 

Of course, this non-universality raises the critical question of how does one choose which heuristic to use? This question closely mirrors the inductive challenge of choosing a prior for a Bayesian model. A general solution to choosing the best heuristic is computationally intractable \cite{Rich2021NaturalismToolbox}, though effective solutions have been offered \cite{rieskamp2006ssl,Scheibehenne2013}. Intuitively, if one believed, for whatever reason, that an environment was governed by numerous additive factors, then a heuristic like TAL would be a good strategy to adopt. The problem of strategy selection closely relates to the problem of meta-learning, or learning to learn, in which one determines how to choose hyperparameters, architectures, general strategies, etc. that will perform well in a task \cite{Schweighofer2003Meta-learningLearning}. With enough data one can test which heuristic performs best on a sub-sample. However, in the low data regime this might not be possible. Our results show the differences between TTB and TAL used as a prior may not be too significant but future work should explore this angle.

With reference to models of human decision making, this class of algorithms has further potential. Referring back to the roots of regularized regression, Tikhonov \cite{tikhonov1943stability} initially constructed this type of regularization in a more general form:

\begin{equation}
\bm{\hat{w}}_{ridge} = \argmin_{\bm{w}} \{ \|\bm{y} - \bm{Xw}\|_2^2  + \|\bm{\Gamma}(\bm{w} - \bm{w}_{prior})\|_2^2\} \label{eq:tikhonov} 
\end{equation}

where $\theta$ has been replaced with a matrix $\bm{\Gamma}$. This enables the implementation of different penalty values for different directions in weight space. Admittedly, choosing $\bm{\Gamma}$ would require knowledge of the data. Our results suggest there might be some advantage in this kind of stepwise approach, where one model's output provides another model's prior. From a psychological point of view, this would enable modelling attention through the scaling of dimensions \cite{nosofsky1986attention}. Although empirical studies show humans usually employ attention solely along individual dimensions (i.e., the diagonal of $\bm{\Gamma}^{T}\bm{\Gamma}$ \cite{Jones2005StimulusLearning,Kruschke1993HumanModels}), other applications (like our fMRI example) could benefit from this generality \cite{Bobadilla-Suarez2020MeasuresSimilarity}. Generalizations of such regression algorithms include adding a matrix that puts weights on observations themselves \cite{van2015lecture} or even using heuristic regularizers for more complex models like neural networks. As in all modeling endeavours, the researcher should make clear how the model is intended \cite{jones2011bayesian}. For instance, a penalized regression approach could be proposed and evaluated as a normative account of what should be done, a high-level description of what people actually do, or an algorithmic account of the processes people engage in. We suggest further work on expertise (e.g., transitioning from novice to expert) could engage with any of the mentioned modelling strategies.

Furthermore, the models presented here provide only point estimates of $\bm{\hat{w}}_{ridge}$, but there is also no obstacle in expanding them to the Bayesian setting to obtain the full posterior distribution as well. In fact, it is well known that ridge regression, like LASSO regression, has a Bayesian interpretation \cite{friedman2001elements,parpart2018heuristics,Tibshirani1996RegressionLasso}. Our two-step approach engages in a double counting of the data (cf. \cite{Zou2006TheProperties}) which could suffer from bias and undue confidence in predictions. A Bayesian formulation could address this potential issue, providing new insights on why our two-step approach works and in which environments. This exciting future direction could expand the reach of our approach by placing it on a normative footing, enabling inquiry into the models' confidence. 

In conclusion, we find that priors motivated by decision heuristics are valuable both methodologically and theoretically. Assuming independence among predictor variables offers a reasonable prior or starting point in most situations. These priors are themselves data-informed models that perform robustly when the penalty value (i.e., prior) is overly strong. Although ridge regression may not routinely suffer from extreme penalty values in practice, use of the TAL and TTB priors do not appear to have any significant downside and may be judged a more sensible choice and perhaps more akin to how people learn than ridge regression's null vector prior. Linking insights across fields as disparate as decision making and advanced methodologies for fMRI data analysis, we are confident that these robust priors for regularized regression will find even further utility in other fields, surpassing the theoretical contributions that we have hinted at here.

\clearpage

\section*{Acknowledgments}
We thank Franziska Bröker, Brett Roads and the rest of the Love Lab for comments on an earlier version of this manuscript. 

\section*{Funding}
This research was supported by the National Institutes of Health (NIH) Grant 1P01HD080679, Royal Society Wolfson Fellowship 183029, and Wellcome Trust Senior Investigator Award WT106931MA from BCL, and National Science Foundation (NSF) Grant 2020906 to MJ.

\section*{Author contributions}
BCL developed the study concept and derived the objective function for Equation \ref{eq:1}. SBS coded all the analyses and simulations, derived the objective function for Equation \ref{eq:wlogridge} along with its Newton-Raphson estimate, and wrote the initial draft. SBS, MJ and BCL interpreted the results and provided critical comments on the manuscript.

\section*{Competing interests}
Authors declare no competing interests.

\section*{Data and materials availability}
\begin{enumerate}
  \item The 20 datasets from Application I can be retrieved from: 
  
  https://osf.io/fg4p5/
  \item The Breast Cancer Wisconsin (Diagnostic) Data Set from Application II can be retrieved from: 
  
  https://archive.ics.uci.edu/ml/datasets/Breast+Cancer+Wisconsin+(Diagnostic)
  
  \item Code and data for Applications I and II can be found at:
  
  https://github.com/bobaseb/robust\_priors\_regression
  \item Code for the simulations of the brain imaging data can be found at a fork of the RSA Toolbox \cite{Nili2014} (develop branch): 
  
  https://github.com/bobaseb/rsa\_toolbox\_lss/tree/develop/LSS\_project
 
\end{enumerate}

\clearpage




\clearpage

\end{document}